# Mulberry Leaf Yield Prediction Using Machine Learning Techniques

Srikantaiah K C[1,*], Deeksha A[1]

[1] Department of Computer Science and Engineering, SJB Institute of Technology, Bangalore, Karnataka, India
* Corresponding author. Email: srikantaiahkc@gmail.com

**ABSTRACT**
Soil nutrients are essential for the growth of healthy crops. India produces a humungous quantity of Mulberry leaves which in turn produces the raw silk. Since the climatic conditions in India is favourable, Mulberry is grown throughout the year. Majority of the farmers hardly pay attention to the nature of soil and abiotic factors due to which leaves become malnutritious and thus when they are consumed by the silkworm, desired quality end-product, raw silk, will not be produced. It is beneficial for the farmers to know the amount of yield that their land can produce so that they can plan in advance. In this paper, different Machine Learning techniques are used in predicting the yield of the Mulberry crops based on the soil parameters. Three advanced machine-learning models are selected and compared, namely, Multiple linear regression, Ridge regression and Random Forest Regression (RF). The experimental results show that Random Forest Regression outperforms other algorithms.

## 1. INTRODUCTION

In India about 97% of raw Mulberry silk is produced due to its salubrious climatic conditions suitable for growing mulberry crops throughout the year. Sericulture encompasses mulberry crops, silkworm and cocoon as its main components and these are interdependent on each other for the production of qualitative quantities of raw silk. Mulberry requires optimum climatic conditions for its growth and development. For feeding silkworm and for the production of biomass, Mulberry leaves with balanced nutrition are essential. Majority of the farmers hardly pay attention to the nature of soil and abiotic factors due to which leaves become malnutritious and thus when they are consumed by the silkworm, desired quality end-product, raw silk, will not be produced.

The factors which affect the growth of healthy Mulberry leaves are soil quality parameters and environmental factors. The soil quality parameters namely, pH, Electrical conductivity (EC ds/m$^2$), Organic carbon (OC%), Average Nitrogen (N Kg/ha), Available Phosphorus (P Kg/ha), Available Potassium (K Kg/ha), Available Sulphur (S ppm), Average Boron (B ppm) and other micronutrients along with environmental factors influence the quality and quantity of Mulberry leaf yield.

Soil pH controls the chemical processes of the soil and is basically an important soil parameter and the availability of all other nutrients in the soil are dependent on the soil pH. Organic Carbon content is also an important parameter, which improves the morpho-physico-chemical properties of soil. The macronutrients responsible for the growth of Mulberry crops are Nitrogen, Phosphorus and Potassium (NPK) and these should necessarily be present in the soil in required levels. Nitrogen is the main component of many structural, genetic (RNA and DNA) and metabolic compounds in the cells and one of the components of chlorophyll, which is used during photosynthesis. Phosphorous, another macronutrient, is required for plant growth and is responsible for energy transfer, photosynthesis and transformation of sugars. Potassium is the next important macronutrient which is present in the vacuole and cytoplasm, without forming organic matter in cells and is responsible for the plant metabolism and helps in prevention of diseases, pests etc.

Other micronutrients such as Sulphur, Boron, Manganese, etc., which are required may be present in the soil, or is given to the plants using foliar spray which is sprayed on the leaves and therefore they enter through the stomata of the leaves. This is because these micronutrients get eroded when present in the soil and are therefore sprayed depending on their availability in the soil.

With suitable environmental conditions along with all these soil parameters together are responsible for the production of qualitative quantities of Mulberry crops. Therefore, using the advanced Machine Learning techniques on the soil data provided by the Karnataka State Sericulture Research and Development Institute, with which we can predict whether the soil in a new place is suitable for good growth of Mulberry crops or not, and hence depending on the soil parameters we can suggest the chemical fertilizers for that particular soil.

To get a good and healthy Mulberry crop yield, usage of suitable fertilizers is important and essential. If the crop yield and quality of the Mulberry crops is not up to the mark, the quality of raw silk, we get will also not be up to the mark. This increases the cost for the farmers and thus they incur loss. It becomes a wild-goose chase for the farmers if the desired quality and quantity of raw silk is not obtained after months of hardwork. Hence, use of fertilizers to improve the soil texture in order to get good quality and quantity of Mulberry crops is required which in turn increases the farmers' profit.

In this paper, we propose a technique to predict the Mulberry leaf yield using Machine Learning models that can be beneficial and profitable to the farmers by giving them an idea about the quantity of leaf yield depending on their soil parameters.

Remainder of the paper is organized as follows - Section 2 is given as a Literature Survey, Section 3 is given as Problem Definition and Methodology, Section 4 is given as Experimental Results and Section 5 concludes the paper and highlights the future scope.

## 2. LITERATURE SURVEY

Ramya *et al.* [1] proposed a qualitative model by analyzing the soil data of Bangalore and Mysore regions of Karnataka containing 10 soil attributes and have predicted a range of soil parameters suitable for the Mulberry crop growth. They have developed a model that classifies the soil as 'IDEAL' or 'NON IDEAL' based on the soil parameters for growth of Mulberry crops making use of Decision tree algorithm using J48 from data mining tool WEKA. Hunt's algorithm was used for generalizing soil parameters.

Singh *et al.* [2] proposed Soil Test based Fertilizers Recommendation of NPK for Mulberry Farming in Acid soils of Lohardaga, Jharkhand. Application of NPK macro nutrients among different farmers' field were categorized into farmers existing practices, recommended practices and soil test-based fertilizer application. The average leaf yield gain was recorded from 10.86 to 15.29 % among the fields. This could be improved using modern technology to directly help the farmers by suggesting them the type and dosage of the fertilizers to be used based on the soil parameters.

Sudhakar et al. [3] research shows that the soils received from the sericulture farming gardens of North, South and Eastern Karnataka showed marked variability and is mostly comprised of clay loamy soils (52%). They analysed a total of 2067 soil samples for their soil reaction (pH), electrical conductivity (EC) and other nutrients and macronutrients. Samples received were air-dried in shade, powdered, passed through a 10µ mesh sieve and stored in a fresh polythene cover with proper labelling. According to their study, they advise to enhance application of the current doses of organic manures, FYM, compost, green manuring for retaining and improving the fertility and health of the soil and to take a routine testing of soil chemicals for at least once or twice in a year.

Ray *et al.* [4] has proposed fertility status of mulberry for mulberry crop production. Importance of essential nutrient elements in soil such as N, P, Zn and B on the quality of both the mulberry leaves and silkworm cocoon is stated. They have used nutrient index (NI) for the assessment of soil fertility. The values of available N, P and K was calculated block wise for surface soil samples using Nutrient index = [(Nl × 1) + (Nm × 2) + (Nh × 3)] / Nt. Data analysis function of SAS software were used for descriptive and correlation analysis. Their study states that the soil and land resources are the most important factors in the production of mulberry crop.

Bordoloi *et al.* [5] have proposed Impact of Soil Health Card Scheme Production, Productivity and Soil Health in Assam. It has covered 120 samples comprising 60 soil health card holders and 60 control farmers which showed that the soil of Assam is acidic in nature. Their study states that for sustainable agriculture, judicious use of fertilizers is must. Otherwise, the farmers will suffer from two possibilities, the first is overdoses of fertilizers which will affect the nature of soil structure and on natural environment too and underdose leads to low productivity and quality of crops.

Sakthivel, *et al.* [6] have proposed Organic Farming in Mulberry which states that organic manures play a vital role on soil health by improving its physical, chemical and biological properties. The two types of methods which were used are physical method which is water jetting and biological method. The biological method plays a significant and indispensable role in managing key pests of mulberry because application of chemical inputs like inorganic fertilizers, weedicides, insecticides, fungicides etc., in bimonthly interval in mulberry gardens not only pollute the ecosystem but also cause adverse impact on the soil health and hazardous effect on human beings and beneficial organisms including silkworms.

Sharma *et al.* [7] have proposed Machine Learning Applications for Precision Agriculture also known as smart farming has developed into an innovative tool for solving current agricultural sustainability challenges. The driving force behind this cutting-edge technology is machine learning (ML). ML together with IoT (Internet of Things) enabled farm machinery are key components of the next agriculture revolution. In this article, the author systematically outlines the application of ML in agriculture. ML is used with computer vision inspection to classify a set of different crop images to monitor crop quality and estimate yield.

Chen *et al.* [8] have proposed "Estimation of Soil Moisture Over Winter Wheat Fields During Growing Season" by making use of machine learning methods with highly nonlinear tuning capabilities not limited to physical parameters. Machine-Learning methods were used over winter wheat fields to estimate soil moisture during its growing season. Around 240 sample plots and RADARSAT-2 data consisting of quad polarization of the study area were acquired and collected. To assess and compare the performance of the model used The root-mean-square error (RMSE) and coefficient of determination (R2) were used.

## 3. PROBLEM DEFINITION AND METHODOLGY

### 3.1 Problem Definition

This paper is aiming at designing an effective Mulberry Leaf yield prediction model to improve the quality and quantity of the Mulberry crops which are fed to the silkworms which in turn produces the raw silk. Machine Learning models, namely Multiple Linear Regression, Ridge Regression and Random Forest Regression are compared and the results are observed [10-11].

### 3.2 Methodology

The architecture for predicting the Mulberry leaf yield consists of the components: Input Dataset, Data Preprocessing, Training the model, Testing the model, Output- Yield prediction. As illustrated in the Figure 1, the model is trained on a dataset consisting of soil attributes. First, it undergoes Data preprocessing, where null values, if any, are removed and an Exploratory data analysis is performed to get useful insights from the dataset. Since the values in different attributes have different ranges of values, all the attributes are Normalized in order to train the model more precisely. If there are Categorical variables, they are encoded into corresponding numeric values since the model requires numeric data to undergo training. The dataset is split into train and test sets and a regression model is applied since the yield of the Mulberry crop has to be determined. Next, appropriate fertilizers will be recommended based on the yield of crops in order to grow good quality Mulberry crops.

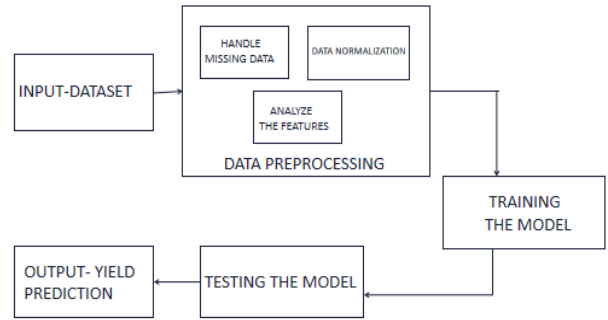

**Figure 1** Mulberry crop yield prediction model

- **Input - Dataset**: Dataset is collected from Karnataka State Sericulture Research and Development Institute (KSSRDI). The dataset consists of soil parameters that influence the growth of Mulberry crops. They are- pH, Electrical Conductivity, Electrical Conductivity, Organic Carbon, Nitrogen, Phosphorous, Potassium, Calcium, Magnesium, Zinc and Iron.

**Table 1.** Soil parameters and their benefits

| ATTRIBUTES | BENEFITS |
|---|---|
| pH | Degree of acidity and alkalinity in soils. The activity of microorganisms, solubility, and availability of nutrients are among the most important pH-dependent processes. |
| Electrical Conductivity (EC) | It is a degree of the quantity of salts in soil (salinity of soil). It is a vital indicator of soil health. |
| Organic Carbon (OC) | It is the crucial factor in preserving soil quality due to its function in enhancing the physical, chemical, and organic properties of the soil. |
| Phosphorus (P) | It is a crucial macro-element required for plant nutrition. It participates in metabolic methods like photosynthesis, energy switch, and synthesis and breakdown of carbohydrates. |
| Potassium (K) | It is a vital plant nutrient and is needed in huge quantities for the proper increase and reproduction of plants. |
| Calcium (Ca) | It is essential for plant growth and makes less susceptible much less at risk of illnesses and pests. |
| Magnesium (Mg) | The main component of chlorophyll. It enables the activation of enzyme systems. |
| Zinc (Zn) | It is an essential trace element for plant growth and plays an important |

|  | role in the catalytic part of various enzymes. |
|---|---|
| Iron (Fe) | Certain enzymatic functions require it. Alkaline soils usually cause iron deficiency, which can be easily solved by adding iron fertilizers. |

- **Data Preprocessing:** Dataset obtained is preprocessed by removing all the null values and irrelevant values. The relationship between the features is analyzed. The dataset consists of wide range of values all of them will be normalized into values ranging between 0 and 1. The goal of normalization is to bring all the values to a common scale.
- **Training the model:** In this stage, the model is being trained on the training set. The dataset is trained using Multiple Linear Regression, Ridge and Random Forest Regression models and the accuracy of these models is compared.

i. Firstly, Multiple linear regression model is used. It is a statistical approach that makes use of numerous explanatory variables to predict the final results of a dependent variable. The relationship of more than one linear regression (MLR) is to find the linear relationship among the independent variables and dependent variables.

ii. Secondly, the dataset is trained using Ridge regression. It is a model tuning approach that is used to research any data that suffers from multicollinearity. This approach performs L2 regularization when the issue results in expected values being some distance far from the real values.

iii. Random Forest regression is an also applied. It is supervised machine learning algorithm that is constructed from decision tree. It utilizes ensemble learning that combines many classifiers to provide solutions to complex problems.

- **Testing the model:** The various regression models that are trained will be tested for the accuracy and the precision. The model will be evaluated using a testing set and the corresponding observations will be recorded.

- **Output - Yield prediction:** The applied model will predict the Mulberry leaf yield based on the soil parameters given as input [12-14]

## 4. EXPERIMENTAL RESULTS

### 4.1 Experimental Setup

The prediction algorithm is implemented using Python. The System requirements include- Processors.1 GB Disk Space, Operating systems like Windows 7 or later, macOS, and Linux. GPU – NVidia TitanX Pascal (12 GB VRAM) GPU instance with AWS/Azure/Google Cloud. The dataset on which the prediction algorithm was obtained is from Karnataka State Sericulture Research and Development Institute. The soil data given by KSSRDI belongs to Dakshina Kannada district. [15-18]

### 4.2 Data Preprocessing

Data Preprocessing is an analysis approach used for summarizing the main characteristics, using statistical graphs and other data visualization methods. The Figure 2 demonstrates the correlation between the leaf yield and the soil nutrient content. The red color demonstrates the positive correlation between the features and gray color demonstrates the negative correlation between the features.

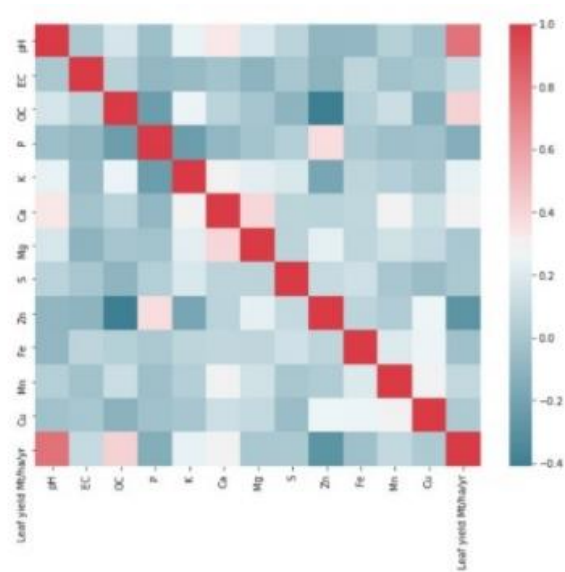

**Figure 2** Heatmap demonstrating the relationship between the various attributes

**Table 2.** Table depicting parameters for sample values

| Parameters | Sample 1 | Sample 2 |
|---|---|---|
| pH | 5.3 | 5.3 |
| EC | 0.16 | 0.09 |
| OC | 0.75 | 0.50 |
| P | 13.0 | 7.0 |
| K | 52.0 | 34.0 |
| Ca | 2.8 | 2.1 |
| Mg | 1.3 | 1.1 |
| S | 19.82 | 11.32 |
| Zn | 2.48 | 2.120 |
| Fe | 13.75 | 15.825 |
| Mn | 38.07 | 10.58 |
| Cu | 1.005 | 0.740 |

**Table 3.** Table depicting results for Sample 1 and 2

| Samples | Actual Yield | MLR Yield | RR Yield | RF Yield |
|---|---|---|---|---|
| Sample 1 | 50.36 | 48.937 | 49.018 | 50.21 |
| Sample 2 | 48.62 | 40.306 | 40.423 | 40.859 |

The accuracy is calculated using accuracy_score function from sklearn library. When using multiple labels for classification, this function calculates the subset accuracy: the predicted label set of the sample must exactly match the corresponding label set in 'y_test'.

The accuracy is obtained by the formula-

$$R^2 = 1 - (RSS/TSS) \qquad (1)$$

where,

$R^2$ = coefficient of determination
RSS = sum of squares of residuals
TSS = total sum of squares

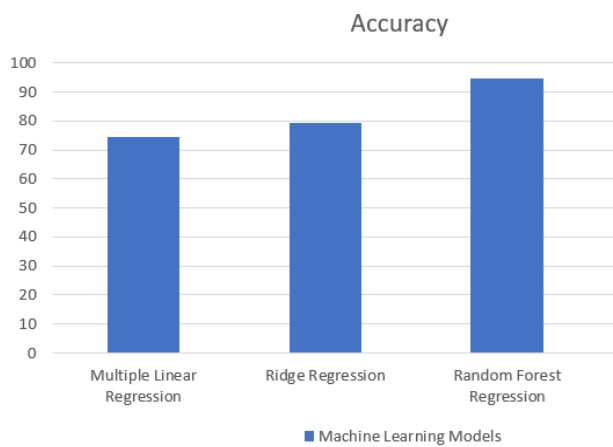

**Figure 4** Comparison of Accuracies of Machine Learning models

It is observed that Random forest Regression has the highest accuracy of 94.6%. Both Ridge and Multiple Linear Regression have 79.4% and 74.31% accuracy respectively. Since the performance of Random forest regression is very exceptional, the same is preferred over the other two [19-23].

## 5. CONCLUSION

In this paper, we have predicted the Mulberry Leaf Yield using the Machine Learning models, namely, Multiple Linear Regression, Ridge Regression and Random forest regression. The accuracies obtained by the three models are compared. The accuracy obtained by the Random Forest Regression model is 94.6%. In the future, the accuracy of the model can be improved using a larger dataset and other feature extraction techniques. As a part of future work, different fertilizers can also be suggested based on the soil paramenters.